\documentclass[letterpaper]{article}
\usepackage{uai2018}
\usepackage[margin=1in]{geometry}

\usepackage{times}
\usepackage{algorithm}
\usepackage{algorithmic}
\usepackage{graphicx}
\usepackage{amsmath}
\usepackage{amssymb}
\usepackage{mathtools}
\usepackage{multirow}
\usepackage{color}
\usepackage{comment}
\usepackage{natbib}

\newtheorem{thm}{Theorem}

\newtheorem{ex}{Example}

\newtheorem{coro}{Corollary}

\newenvironment{proof}{\noindent{\bf {Proof: }}\ }{\hfill$\blacksquare$ \vspace{1mm}}

\newcommand{\prr}[1]{\Pr\nolimits_{r}(#1)}

\title{A Cost-Effective Framework for Preference Elicitation and Aggregation}

\author{{\bf Zhibing Zhao, Haoming Li, } \\
{\bf Junming Wang}\\
RPI\\
Troy, NY, USA\\
\{zhaoz6,lih14,wangj33\}@rpi.edu\\
\And
{\bf Jeffrey O.~Kephart,} \\
{\bf  Nicholas Mattei, Hui Su}\\
IBM Research\\
Yorktown, NY, USA\\
\hspace{15mm}\{kephart,n.mattei,huisuibmres\}@us.ibm.com\\
\And
{\bf Lirong Xia} \\
RPI\\
Troy, NY, USA\\
xial@cs.rpi.edu\\
}

\begin{document}

\maketitle

\begin{abstract}
We propose a cost-effective framework for preference elicitation and aggregation under the Plackett-Luce model with features. Given a budget, our framework iteratively computes the most cost-effective elicitation questions in order to help the agents make a better group decision. 

We illustrate the viability of the framework with experiments on Amazon Mechanical Turk, which we use to estimate the cost of answering different types of elicitation questions.  
We compare the prediction accuracy of our framework when adopting various information criteria that evaluate the expected information gain from a question. Our experiments show carefully designed information criteria are much more efficient, i.e., they arrive at the correct answer using fewer queries, than randomly asking questions given the budget constraint.
\end{abstract}

\section{INTRODUCTION}\label{sec:introduction}

Consider the hiring decision problem~\citep{Bhattacharjya14:Bayesian}. With the aid of an intelligent system, a group of people (the {\em key group}) faces a hiring decision about many candidates who are characterized by attributes, such as experiences, technical skills, communication skills, etc. The goal is to help the key group make a group decision without directly eliciting their full preferences over all candidates, which is often infeasible given the vast number of candidates. Instead, the intelligent system may ask fellow employees (the {\em regular group}) about their preferences in order to learn about the key group's preferences. 
How can the intelligent system decide which member in the regular group to ask and which questions to ask?  Note that we discuss the presence of two groups but our framework is applicable when there is only one group of decision makers as well.

This example illustrates the {\em preference elicitation} problem, which has been widely studied in the field of recommender systems~\citep{Loepp14:Choice}, 
healthcare~\citep{Chajewska00:Making,Weernink14:Systematic, Erdem17:Preferences}, 
marketing~\citep{Huang16:Consumer}, stable matching~\citep{Drummond14:Preference,Rastegari16:Preference}, combinatorial auctions~\citep{Sandholm06:Preference},
 etc. Most previous works studied a special case of the aforementioned scenario, in which the regular group is the key group. The objective of preference elicitation is to achieve some goal using as few samples (data) as possible. A common approach is to adaptively ask questions that maximize expected information gain, measured by some information criteria. 


Moreover, most previous work focused on specific types of elicitation questions, e.g. pairwise comparisons. In this paper, we consider a more general framework that asks a variety of elicitation questions and can accommodate one or more groups. 
The diversity of elicitation questions enables us to query cost-effectively. Intuitively, an agent's preference order over 10 alternatives tells us more about her preference in general than just her top choice among the 10; however, it may take her longer to do so. The key question we want to answer in this paper is: 

{\em How can we compute the most cost-effective questions for preference elicitation under resource constraints?}

\subsection{OUR CONTRIBUTIONS} 

We propose a flexible cost-effective preference elicitation and aggregation framework to predict a single agent's preference or help make a group decision. The main inputs include a budget $W$, a set of designs (i.e.~questions to ask) $\mathcal H$, a cost function $w$,  a randomized voting rule and an information criterion. We model non-deterministic preferences using the Plackett-Luce model with features.

\begin{figure}[htp]
	\centering
	\includegraphics[width=0.45\textwidth]{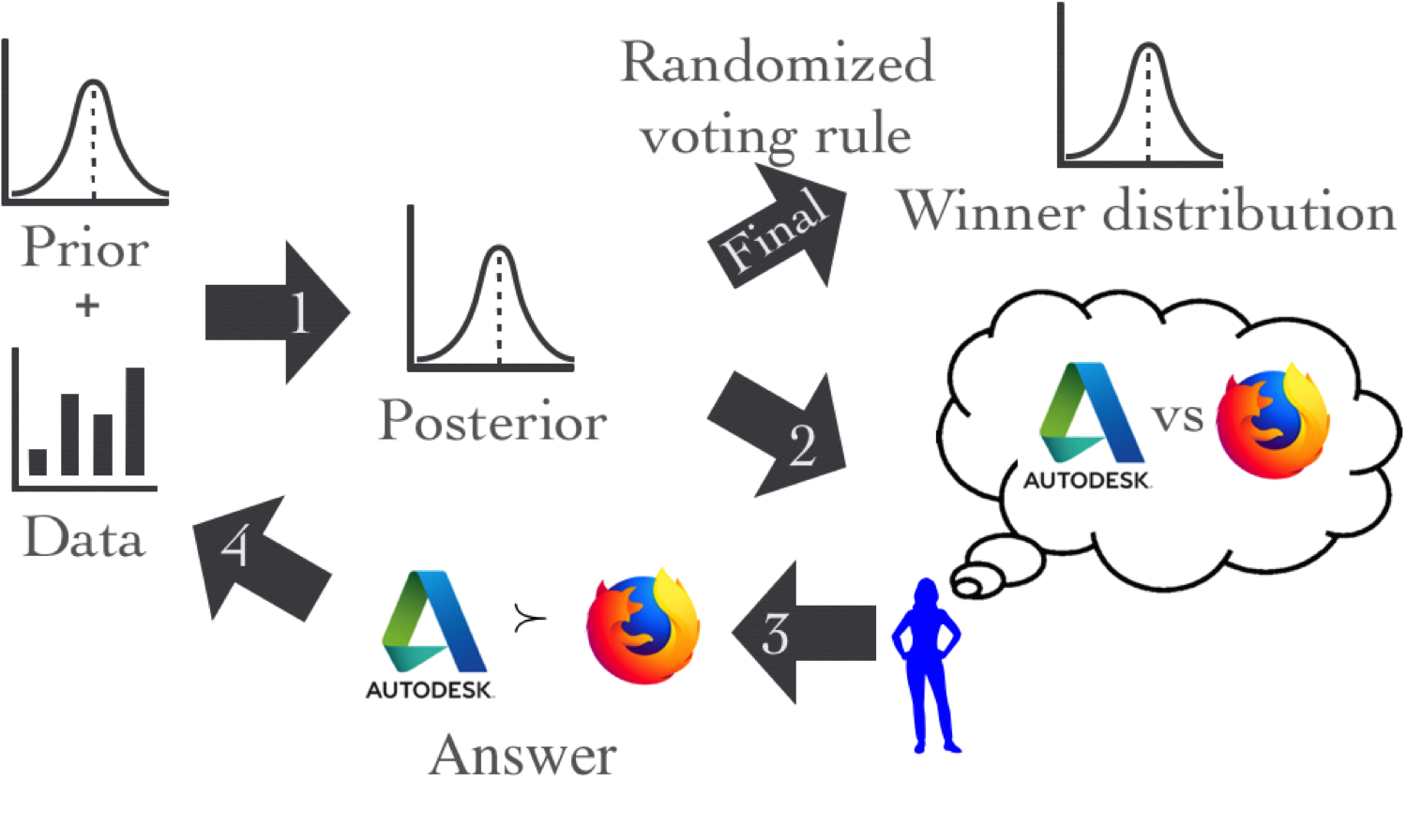}
	\caption{Illustration of the proposed framework.}
	\label{fig:procedure}
\end{figure}

{\bf Cost-effectiveness.} We propose a flexible, cost-effective preference elicitation framework that accommodates all randomized voting rules, ranking models, and information criteria. This iterative framework leverages the {\em optimal design} technique. In each iteration, we choose the question that provides the most information per unit cost. 
The response is then recorded as a data point, leading to an update of the posterior distribution of the parameter, which is treated as the prior for the next iteration. In any iteration, the posterior estimate of the parameter can be used to compute a winner distribution using a randomized voting rule. This procedure is illustrated in Figure~\ref{fig:procedure}.

{\bf Randomized Voting Rules.} 
We use randomized voting rules to compute the winning alternatives of a group decision, which outputs the distribution of winners (see Section~\ref{sec:rvr} for details). The probability for each alternative to be the winner is proportional to its score based on the voting rule. 
These probability estimates are more informative than only recommending a winner as it provides a distribution over the candidates as well.

We prove that when people have non-deterministic preferences, the probability of an alternative to be the winner is proportional to the total expected score of this alternative for all agents (Theorem~\ref{thm:general}). This means the randomized counterpart of any scoring rule can be used in our framework as long as the expected score of each alternative for a single agent is easy to compute. Then we prove that under the Plackett-Luce model, the winner distributions of probabilistic plurality and probabilistic Borda are easy to compute (Corollary~\ref{coro:plu} and Theorem~\ref{thm:borda}). 

{\bf Information Criteria.} An information criterion plays a key role in determining the next elicitation question by measuring the information in the distribution of a parameter. We propose the minimum pairwise certainty (MPC) criterion, extended from the information criterion by~\cite{Azari13:Preference}, which maximizes the improvement of the least certain pairwise comparison. Other commonly-used information criteria include D-optimality~\citep{Wald43:Efficient,Mood46:Hotelling} and E-optimality~\citep{Ehrenfeld55:Efficiency}, as well as asking a question uniformly at random. All these information criteria are based on the information of the posterior distribution of the model parameter, which is approximated by its asymptotic distribution, a multivariate Gaussian computed based on the composite marginal likelihood method~\citep{Pauli11:Bayesian}.

{\bf Empirical Studies \& Experiments.} We carry out Amazon Mechanical Turk experiments to estimate the cost of answering various types of questions for a target domain of ranking hotels. We compare the performances of MPC, D-optimality and E-optimality with simulations and observe that these criteria have similar performance in terms of prediction accuracy, and we observe that all of them significantly outperform random elicitation questions. 



\subsection{RELATED WORK AND DISCUSSIONS} 

Our work is related to cost-effective experimental designs, which were investigated by \cite{Wright10:Workweek,Volkov14:Optimal} in the context of aquatic toxicology and drug development, respectively. \cite{Volkov14:Optimal} modeled cost-effectiveness as different types of optimization problems, e.g., minimize cost under information constraints. We take a greedy approach, similar to an algorithm proposed by \cite{Wright10:Workweek}, and choose the design (elicitation question) that maximizes the expected information gain per unit cost. Our cost varies depending on the type of questions and is estimated empirically, similar to the idea in \cite{Volkov14:Optimal}. To our best knowledge, this paper is the first work to apply cost-effective experimental design to preference elicitation. 

The greedy approach is also called one-step-lookahead policy, which can be arbitrarily worse than optimal $t$-step-lookahead ($t$-step myopic active search) policies for $t\ge 2$~\citep{Garnett12:Bayesian}. Arbitrary $t$-step myopic active search is hard to compute, as was shown by \cite{Jiang17:Efficient}, which also proved that nonmyopic active search is computationally hard even to approximate and proposed an efficient searching algorithm. This algorithm is potentially useful in the preference elicitation context and is an interesting future direction.

Most previous works in preference elicitation assumed that people's preferences are deterministic. For example, \cite{Bhattacharjya14:Bayesian} proposed an even swap algorithm to reveal a single decision maker's most preferred alternative; \cite{Lu11:Robust,Kalech11:Practical} elicited preferences from a group of people in order to make a group decision under a (deterministic) voting rule. In contrast, we consider non-deterministic preferences of people, which is often the case in real-world. Moreover, we use randomized voting rules, which output the probability of each alternative to be the winner. These probabilities, which can be viewed as normalized scores over all alternatives, provide a quantitative measure of the quality of each alternative. For example, an alternative that wins with probability 0.8 can be seen as being much better than other alternatives.

Non-deterministic preferences were modeled by general random utility models by~\cite{Azari13:Preference}. They proposed a preference elicitation framework for personalized choice and social choice (aggregated preference). 
We use the Plackett-Luce model with features, which is a special case of general random utility models but has easy-to-compute probabilities. More importantly, we use randomized voting rules for aggregation, which is very different from parametric modeling of social choices employed by~\citep{Azari13:Preference}.

Pairwise elicitation questions may be the most widely explored in the literature due to their simplicity~\citep{Branke17:Efficient,Eric08:Active,Houlsby12:Collaborative,Lu11:Robust,Pfeiffer12:Adaptive}. In contrast, \cite{Azari13:Preference} focused on elicitation of full rankings, though their proposed framework also allows for partial orders. \cite{Drummond14:Preference, Lu11:Vote} studied a larger set of queries, which includes asking a person to rank her top $k$ choices over all alternatives. In this paper, we consider an even broader set of queries, asking an agent to rank her top $k$ choices over a subset of $l$ alternatives ($k < l$). This enables us to elicit preferences in a more cost-effective manner. 

As a key role in preference elicitation, information criteria have been widely investigated for different applications. Standard information criteria include D-optimality (used in~\citep{Houlsby11:Bayesian,Houlsby12:Collaborative,Pfeiffer12:Adaptive}) and E-optimality. \cite{Drummond14:Preference} and~\cite{Lu11:Robust} use minimax-regret-based criterion for stable matching and aggregation respectively. \cite{Azari13:Preference} proposed yet another criterion, defined on the certainty of the least certain pairwise comparison over the intrinsic utilities (part of the parameter of their general random utility models) of all alternatives. Our MPC criterion extends the criterion by \cite{Azari13:Preference}. To predict a single agent's top $k$ preference, we search over a subset of all pairwise comparisons (see Section~\ref{sec:mpc}). To help make a group decision, we search over all pairwise comparisons of all agents in the key group to find the least certain pairwise comparison (Equation~\eqref{eq:mpc}). 



\section{PRELIMINARIES}

Let $\mathcal A=\{a_1, a_2, \ldots, a_m\}$ denote a set of $m$ alternatives and $\{1, \ldots, n_1, n_1+1, \ldots, n_1+n_2\}$ denote $n_1+n_2$ agents, where the first $n_1$ agents belong to the key group, who will be making a group decision. The remaining $n_2$ agents belong to the regular group. For all $i=1, \ldots, m$, $a_i$ is characterized by a real-valued column vector of $K$ attributes $\vec z_i$. For all $j=1, \ldots, n_1+n_2$, agent $j$ is characterized by a real-valued column vector of $L$ attributes $\vec x_j$. A full ranking $R$ is often denoted by $a_{i_1}\succ a_{i_2}\succ\ldots\succ a_{i_m}$, where ``$\succ$" means ``is preferred over". We denote the budget by $W$, where the money is used to pay the agents for answering elicitation questions.

For $n_1=1$, we want to predict the single key agent's full or top $k$ ranking with as much certainty as possible given a budget $W$. For $n_1\ge 2$, the goal is to predict the winning alternative of the key group by eliciting preferences from the regular group in the most cost-effective way. More concretely, given $W$, we want to output a distribution of winning alternatives, w.r.t. a randomized voting rule, which will be defined in Section~\ref{sec:rvr}. 

\subsection{THE PLACKETT-LUCE MODEL WITH FEATURES}\label{sec:pl}

Let the parameter $B=[b_{\kappa\iota}]_{K\times L}$ be a matrix of real-valued coefficients, transforming features to utilities. Each value $b_{\kappa\iota}$ corresponds to the $\kappa$-th attribute from an alternative and $\iota$-th attribute from an agent. The parameter space $\Theta$ is a set of all real-valued $K\times L$ matrices.
Then the utility of an alternative $a_i$ to an agent $j$ is 
\begin{equation}\label{eq:utility}
u_{ji}=\vec x^\top_j B\vec z_i.
\end{equation}
For any agent $j$ and any full ranking $R_j=a_{i_1}\succ a_{i_2}\succ\ldots\succ a_{i_m}$, the probability of $R_j$ is
\begin{align*}
\Pr(R_j)&=\frac {\exp(u_{ji_1})} {\sum^m_{q=1} \exp(u_{ji_q})}\times\frac {\exp(u_{ji_2})} {\sum^m_{q=2} \exp(u_{ji_q})}\times\\
&\cdots\times\frac {\exp(u_{ji_{m-1}})} {\exp(u_{ji_{m-1}})+\exp(u_{ji_m})}.
\end{align*}
Given the Plackett-Luce model with features, the probability of alternative $a_{i_1}$ to be ranked at the top among $\{a_{i_1}, \ldots, a_{i_{l}}\}$ by agent $j$ is $\frac {\exp(u_{ji_1})} {\sum^{l}_{q=1} \exp(u_{ji_q})}$. Specifically, for any two alternatives $a_1$ and $a_2$, the probability of $a_1\succ a_2$ by agent $j$ is $\frac {\exp (u_{j1})} {\exp (u_{j1})+\exp (u_{j2})}$.

%

\subsection{ONE-STEP BAYESIAN EXPERIMENTAL DESIGN}

Given any probabilistic model parameterized by $B\in\Theta$ and any prior distribution $\pi(B)$, a one-step Bayesian experimental design consists of two parts: (i) a set of designs $\mathcal H$, where each $h\in\mathcal H$ is composed of an agent and a question; (ii) an information measure $G(\cdot)$, which maps any distribution of $B$ over $\Theta$ to a real-valued scalar: a measure of information in this distribution. 

For any design $h\in\mathcal H$, the distribution of responses can be computed using the ground truth parameter $B^*$. We use $D$ to denote the set of all possible responses. Given a ground truth parameter $B^*$, the probability of any data $d\in D$ can be computed as $\Pr(d|h)$. Further, we can compute the posterior distribution of parameter $\pi(B|d, h)$ over the parameter space $\Theta$ and the corresponding information criterion $G(\pi(B|d, h))$. The expected information is
$$E[G(\pi(B|h))]=\sum_{d\in D} G(\pi(B|d, h))\Pr(d|h),$$
where the expectation is taken over all possible responses. The goal is to find the design $h$ that maximize the expected information gain, which is $E[G(\pi(B|h)]-G(\pi(B))$, per unit cost. Let $w(h)$ denote the cost function, which maps the 2-tuple (agent, question) to a positive cost. 
Given the cost function $w(h)$, we can compute the optimal design $h^*$ that maximizes the expected information gain per unit cost by
\begin{equation}\label{eq:hstar}
h^*=\arg\max_h\frac {E[G(\pi(B|h))]-G(\pi(B))} {w(h)}.
\end{equation}

\section{COST-EFFECTIVE PREFERENCE ELICITATION}

In our proposed framework, we iteratively adapt the one-step experimental design by querying the most cost-effective question in each iteration. At any iteration $t$, the prior distribution of $B$ is the posterior distribution given data $D^t$, i.e. $\pi(B^t|D^t)$. Given this posterior, we find the most cost-effective design $h^t$, which consists of one agent and one question, and query $h^t$. The response is combined with $D^t$ to form $D^{t+1}$. Then the budget $W$ and the set of designs $\mathcal H$ are updated before going to the next iteration. Finally, when $n_1=1$, we compute the predicted preference of this agent; when $n_1\ge 2$, we compute the distribution of winners based on a randomized voting rule. This framework is formally illustrated in Algorithm~\ref{alg:framework}.

\begin{algorithm}
{\bf Input}: Budget $W$, randomize voting rule $r$, cost function $w(h)$, information criterion $G(\pi(B))$, the set of designs $\mathcal H$ where for any $h\in\mathcal H$, $w(h) \le W$.\\
{\bf Output}: A predicted preference when $n_1=1$ or a distribution of winning alternatives for group decision when $n_1\ge 2$.\\
{\bf Initialization}: Randomly initialize data $D^1$.

\begin{algorithmic}
\WHILE{$\mathcal H$ is not empty}
\STATE Compute/approximate $\pi(B^t|D^t)$;
\STATE Compute $h^t\in\mathcal H$ using \eqref{eq:hstar};
\STATE Implement $h^t$ (query an agent a question). Let $R^t$ denote her answer. Then $D^{t+1}\leftarrow D^t\cup\{R^t\}$, $\mathcal H\leftarrow\mathcal H-h^t$, $W\leftarrow W-w^t$;
\STATE Remove all $h'$'s from $\mathcal H$ where $w(h') > W$.
\ENDWHILE
\STATE Compute the predicted preference when $n_1=1$ or a distribution of winning alternatives according to the voting rule $r$ when $n_1\ge 2$.
\end{algorithmic}
\caption{Cost-Effective Preference Elicitation}
\label{alg:framework}
\end{algorithm}

For the rest of this section, we will explain how to approximate the posterior distribution $\pi(B^t|D^t)$ and how $G(\pi(B))$ is computed. 

\subsection{APPROXIMATION OF POSTERIOR DISTRIBUTION}

For any prior $\pi(B)$ and data $D$, the posterior distribution is given by
$\pi(B|D)=\frac {\Pr(D|B)\pi(B)} {\int_\Theta\Pr(D|B)\pi(B)dB}$ according to Bayes' rule.
This posterior is often hard to compute. A commonly-used approach is to approximate it by its asymptotic distribution, which is a multivariate Gaussian distribution characterized by the composite marginal likelihood (CML) method~\citep{Pauli11:Bayesian}. 

For convenience we vectorize $B$ as a column vector, denoted by $\vec\beta=\text{vec}(B)$. The composite marginal likelihood method~\citep{Lindsay88:Composite,Zhao18:Composite} computes the estimate of the ground truth parameter from marginal events, e.g., pairwise comparisons. Let $\{\mathcal E_1, \ldots, \mathcal E_q\}$ denote $q$ selected marginal events. Then the composite marginal likelihood method computes the estimate $\vec\beta_{\text{CML}}$ by
$$\vec\beta_{\text{CML}}=\arg\max_{\vec\beta\in\Theta}\text{CLL}(\vec\beta)=\arg\max_{\vec\beta\in\Theta}\sum^q_{\lambda=1}\ln\Pr(\mathcal E_\lambda|\vec\beta),$$
where CLL$(\vec\beta)$ denotes the composite log-likelihood function. Under our Plackett-Luce model with features, CLL$(\vec\beta)$ is twice differentiable for all $\vec\beta\in\Theta$, i.e. $J(\vec\beta) = -\nabla^2_{\vec\beta}\text{CLL}(\vec\beta)$ exists. From~\cite{Pauli11:Bayesian}, asymptotically, $\pi(\vec\beta |D)$ is a multivariate Gaussian distribution, whose mean is $\vec\beta_{\text{CML}}$ and covariance matrix is $J^{-1}(\vec\beta)$.

Computing $J(\vec\beta)$ requires computation of second order partial derivatives of $\ln\Pr(\mathcal E_\lambda|\vec\beta)$ for all $\lambda$. We will show the close-form second order partial derivative formula for any response from an agent.

\subsection{THE SET OF DESIGNS} 

Each design $h\in\mathcal H$ is a combination of an agent and a question about her preferences. The agent can be anyone from $\{1, \ldots, n_1+n_2\}$. In this paper, for simplicity, we consider the case where only the agents from the regular group $\{n_1+1, \ldots, n_1+n_2\}$ are queried. 
For any integers $k<l\le m$, we may ask an agent to rank her top $k$ alternatives over a subset of $l$ alternatives. When $k=1, l=2$, the question is a pairwise comparison; when $k=1, l>2$, the question is to query an agent's top alternative among a subset of alternatives; when $k=l-1$, we are asking a full ranking over a subset of alternatives. The advantage of this type of questions is that the probabilities of responses of these questions are easy to compute, as well as their partial derivatives. W.l.o.g. let $R_j=a_1\succ a_2\succ\ldots\succ a_{k}\succ\text{others}$ be the answer from agent $j$. Then we have $\Pr(R_j|B)=\prod^{k}_{p=1}\frac {\exp(u_{jp})} {\sum^{l}_{i=p}\exp(u_{ji})}$, and
$\ln\Pr(R_j|B)=\sum^{k}_{p=1}(u_{jp}-\ln\sum^{l}_{i=p}\exp(u_{ji})).$

For any $1\le\kappa\le K$ and $1\le\iota\le L$, let $b_{\kappa\iota}$ be the $(\kappa, \iota)$ entry of $B$. We have
$$\frac {\partial\ln\Pr(R_j|B)} {\partial b_{\kappa\iota}}=\sum^{k}_{p=1}(\frac {\partial u_{jp}} {\partial b_{\kappa\iota}}-\frac {\sum^{l}_{i=p}\exp(u_{ji})\frac {\partial u_{ji}} {\partial b_{\kappa\iota}}} {\sum^{l}_{i=p}\exp(u_{jp})}),$$
where $\frac {\partial u_{jp}} {\partial b_{\kappa\iota}}$ and $\frac {\partial u_{ji}} {\partial b_{\kappa\iota}}$ are constants (products of an agent's attribute and an alternative's attribute) by definition. Therefore, for diagonal entries, the second order partial derivatives are given by
\begin{align*}
\frac {\partial^2\ln\Pr\nolimits_{j}(R|B)} {\partial b^2_{\kappa\iota}}&=\sum^{k}_{p=1}((\frac {\sum^{l}_{i=p}\exp(u_{ji})\frac {\partial u_{ji}} {\partial b_{\kappa\iota}}} {\sum^{l}_{i=p}\exp(u_{ji})})^2\\
&-\frac {\sum^{l}_{i=p}\exp(u_{ji})(\frac {\partial u_{jp}} {\partial b_{kl}})^2} {\sum^l_{p=1}\exp(u_{jp})}),
\end{align*}
and for non-diagonal entries, we have
\begin{align*}
&\frac {\partial^2\ln\Pr\nolimits_{j}(R|B)} {\partial b_{\kappa_1\iota_1}\partial b_{\kappa_2\iota_2}}\\
=&\sum^{k}_{p=1}(\frac {(\sum^{l}_{i=p}\exp(u_{ji})\frac {\partial u_{ji}} {\partial b_{\kappa_1\iota_1}})(\sum^{l}_{i=p}\exp(u_{ji})\frac {\partial u_{ji}} {\partial b_{\kappa_2\iota_2}})} {(\sum^{m'}_{i=p}e^{u_{ji}})^2}\\
-&\frac {\sum^{l}_{i=p}\exp{u_{ji}}(\frac {\partial u_{ji}} {\partial b_{\kappa_1\iota_1}})(\frac {\partial u_{ji}} {\partial b_{\kappa_2\iota_2}})} {\sum^{l}_{i=p}\exp(u_{ji})}).
\end{align*}


\subsection{INFORMATION CRITERIA}\label{sec:mpc}

An information criterion maps the distribution of a parameter to a real-valued quality. Standard information criteria are mostly directly computed from the covariance matrix $J^{-1}(\vec\beta)$ or its inverse $J(\vec\beta)$. For example, D-optimality~\citep{Wald43:Efficient,Mood46:Hotelling} computes the determinant of $J(\vec\beta)$; E-optimality~\citep{Ehrenfeld55:Efficiency} computes the minimum eigenvalue of $J(\vec\beta)$. We propose the following minimum pairwise certainty (MPC) criterion by extending the criterion from~\citep{Azari13:Preference} to our domain. 

\noindent{\bf MPC for Case $n_1=1$.} We consider two types of purposes: predicting the agent's (unordered) top $k$ alternatives and predicting the agent's ranked top $k$ alternatives. We note that the criterion by~\cite{Azari13:Preference} only applies to full rankings, which is a special case of our ranked top $k$. The intuition of this criterion is to maximize the certainty of the least certain pairwise comparison among a subset of pairwise comparisons. Formally, let $\mathcal A_k$ denote the set of predicted top $k$ alternatives for this key agent.

$\bullet$ Unordered top-$k$ where $1\le k < m$: $$G(\pi(\vec\beta))=\min_{i_1\in\mathcal A_k, i_2\not\in\mathcal A_k}\frac {|\text{mean}(u_{1i_1}-u_{1i_2})|} {\text{std}(u_{1i_1}-u_{1i_2})}.$$
$\bullet$ Ranked top-$k$ where $1 < k < m$: 
$$G(\pi(\vec\beta))=\min_{i_1\in\mathcal A_k, i_2\neq i_1}\frac {|\text{mean}(u_{1i_1}-u_{1i_2})|} {\text{std}(u_{1i_1}-u_{1i_2})}.$$
In the above equations, $\text{mean}(u_{ji_1}-u_{ji_2})$ is computed using $\vec\beta_\text{CML}$ and $\text{std}(u_{ji_1}-u_{ji_2})$ is computed using the approximated covariance matrix $J^{-1}(\vec\beta)$ as follows. 

Because $u_{ji_1}-u_{ji_2}$ is linear with $\vec\beta$ (see Equation~\eqref{eq:utility} and recall that $\vec\beta$ is the vectorization of $B$), we write it as $u_{ji_1}-u_{ji_2}=\sum_{\kappa, \iota}c_{\kappa\iota}b_{\kappa\iota}$, where $c_{\kappa\iota}$'s are constants computed from attributes of $a_{i_1}, a_{i_2}$ and agent $j$. Then we have
$\text{std}(u_{ji_1}-u_{ji_2})=\sqrt{\sum_{(\kappa_1, \iota_1), (\kappa_2, \iota_2)}c_{\kappa_1\iota_1}c_{\kappa_2\iota_2}\text{Cov}(b_{\kappa_1, \iota_1}, b_{\kappa_2, \iota_2})}$. 
When $\kappa_1=\kappa_2=\kappa$ and $\iota_1=\iota_2=\iota$, $\text{Cov}(b_{\kappa_1, \iota_1}, b_{\kappa_2, \iota_2})$ reduces to $\text{Var}(b_{\kappa\iota})$. Both $\text{Cov}(b_{\kappa_1, \iota_1}, b_{\kappa_2, \iota_2})$ and $\text{Var}(b_{\kappa\iota})$ are entries of $J^{-1}(\vec\beta)$.

\noindent{\bf MPC for Case $n_1\ge 2$.} Our MPC for this case is different from the criterion by~\cite{Azari13:Preference} in that we find the least certain pairwise comparison across all agents in the key group. Formally, our MPC for $n_1\ge 2$ is
\begin{equation}\label{eq:mpc}
G(\pi(\vec\beta))=\min_{j\in\{1, \ldots, n_1\}, i_2\neq i_1}\frac {|\text{mean}(u_{ji_1}-u_{ji_2})|} {\text{std}(u_{ji_1}-u_{ji_2})},
\end{equation}
where the computation of $\text{mean}(u_{ji_1}-u_{ji_2})$ and $\text{std}(u_{ji_1}-u_{ji_2})$ are similar to the $n_1=1$ case.

\section{RANDOMIZED VOTING RULES}\label{sec:rvr}

We use randomized voting rules to aggregate the key group's preferences. A randomized voting rule computes the distribution of winners given the preferences of the agents from the key group. Under non-deterministic preferences, this distribution can be computed from the parameter of the model. This section shows that under the Plackett-Luce model with features, probabilistic plurality and probabilistic Borda are easy to compute.

A randomized voting rule assigns a probability for each alternative to be the winner according to the data, usually based on a scoring function. For example, the probabilistic plurality rule, which is equivalent to {random dictatorship}~\citep{Gibbard77:Manipulation}, samples a winner from a distribution where the probability of each alternative being the winner is proportional to the plurality score of this alternative. Other randomized voting rules can be defined similarly, including probabilistic Borda~\citep{Heckelman03:Probabilistic}. The voting rule must have scores associated with it, but this is a very mild restriction because many commonly-studied voting rules including all positional scoring rules, Copeland, range voting, and approval voting, have randomized counterparts. 

Recall that $n_1$ agents are making a group decision among $m$ alternatives. Let $P$ denote the preference profile that consists of $n_1$ full rankings over $m$ alternatives from the key group. Let $s_r(a_i, P)$ denote the score of alternative $a_i$ under voting rule $r$ and $\prr{a_i|P}$ be the probability for $a_i$ to win under the randomized analogy of $r$ given $P$. Then $\prr{a_i|P}$ is computed by
$\prr{a_i|P}=\frac {s_r(a_i, P)} {\sum^m_{i=1}s_r(a_i, P)}.$

\begin{ex}\label{ex:profile}
Suppose the set of alternatives is $\{a_1, a_2, a_3\}$ and the votes are $\{a_1\succ a_2\succ a_3, a_1\succ a_3\succ a_2, a_2\succ a_1\succ a_3\}$. The plurality and Borda scores are shown in Table~\ref{tab:example}. Under probabilistic plurality rule, $a_1$ wins with probability $2/3$ and $a_2$ wins with probability $1/3$. Under probabilistic Borda, $a_1, a_2, a_3$ win with probabilities $5/9, 3/9, 1/9$ respectively. 
\begin{table}[htp]
\begin{center}
\begin{tabular}{|c|c|c|c|}
\hline
& $a_1$ & $a_2$ & $a_3$\\
\hline
plurality & 2 & 1 & 0\\
\hline
Borda & 5 & 3 & 1\\
\hline
\end{tabular}
\end{center}
\caption{Scores under plurality and Borda}
\label{tab:example}
\end{table}%
\end{ex}

We consider non-deterministic preferences from agents, where the preferences from the key agents are independent of each other. Because each agent has $m!$ possible rankings, there are $(m!)^{n_1}$ possible preference profiles. Then we have
$\prr{a_i}=\sum^{(m!)^{n_1}}_{q=1}\Pr(P_q)\prr{a_i|P_q}$,
where $P_q$ denotes the $q$-th possible preference profile. 

Given a voting rule $r$, let $X_{ji}$ be the score of $a_i$ for agent $j$. $X_{ji}$ is a random variable due to the uncertainty of agent $j$'s preference. The following theorem shows that the probability of $a_i$ being the winner is proportional to the sum of expected score of $a_i$ for each agent. 

\begin{thm}\label{thm:general} For any $1\le i\le m$, $\prr{a_i}\propto\sum^{n_1}_j EX_{ji}$.
\end{thm}
\begin{proof} It suffices to prove $\prr{a_1}\propto\sum^{n_1}_j EX_{j1}$. 

By definition, $\prr{a_1}=\sum^{(m!)^{n_1}}_{q=1}\Pr(P_q)\prr{a_1|P_q}$. Let $S$ denote the score of $a_1$ under rule $r$. Then $S$ is a random variable defined over the $(m!)^{n_1}$ cases. Let $s_q$ denote the value that $S$ takes for case $q$. In any case $q$, we have $\prr{a_1|P_q}\propto s_q$ by the definition of randomized voting rules. We re-write it as $\prr{a_1|P_q}=\frac {s_q} {M}$, where $M$ is the normalization factor. Observe that across all the $(m!)^{n_1}$ cases, $M$ does not change because  the voting rule $r$ and the set of agents does not change. So we have
\begin{align}\label{eq:pra1}
\prr{a_1}&=\frac {\sum^{(m!)^{n_1}}_{q=1}\Pr(P_q)s_q} M\notag\\
&\propto\sum^{(m!)^{n_1}}_{q=1}\Pr(P_q)s_q=ES.
\end{align}
Since $S=\sum^{n_1}_{j=1}X_{j1}$, due to linearity of expectation, we have
$ES=E[\sum^{n_1}_{j=1}X_{j1}] = \sum^{n_1}_j EX_{j1}$. By \eqref{eq:pra1}, we have $\prr{a_1}\propto\sum^{n_1}_j EX_{j1}$.
\end{proof}

For probabilistic plurality, the expected score of $a_i$ for agent $j$ is exactly the probability of $a_i$ being ranked at the top by agent $j$. So we have the following corollary:
\begin{coro}\label{coro:plu} Let $p^{a_i}_j$ be the probability of $a_i$ being ranked at the top by agent $j$. For any $1\le i\le m$, $\Pr_{\text{plurality}}(a_i)=\frac 1 {n_1}\sum^{n_1}_j p^{a_i}_j$.
\end{coro}

\begin{thm}\label{thm:borda}
Let $p_j^{a_{i}\succ a_{i'}}$ denote the probability for agent $j$ to prefer alternative $a_{i}$ over $a_{i'}$. Then for any $1\le i\le m$, $\Pr_{\text{Borda}}(a_i)\propto\sum^{n_1}_j\sum_{i'\ne i}p^{a_i\succ a_{i'}}_j$.
\end{thm}
\begin{proof}
For any $i\in\{1, \ldots, m\}$, we have $\Pr_{\text{Borda}}(a_i)\propto\sum^{n_1}_j EX_{ji}$ by Theorem~\ref{thm:general}, where $X_{ji}$ here denotes the score of $a_i$ for agent $j$ under Borda. We only need to prove $EX_{ji}=\sum_{i'\ne i}p^{a_i\succ a_{i'}}_j$. This is a known result, but we were not able to find a formal proof in literature, except a proof for three alternatives by~\cite{Chen05:Winning}, which is easy to be extended for arbitrary number of alternatives. For completeness we provide a short proof.

By definition of Borda, we have $EX_{ji}=\sum^{m-1}_{k=1}(m-k)\sum_{R: a_i\text{ at $k$th position of }R}\Pr_j(R)$, where $R$ is any full ranking over the $m$ alternatives and $\Pr_j(R)$ is the probability of $R$ by agent $j$. Imagine $m-1$ bins, each of which is labeled with $a_i\succ a_{i'}$ for all the remaining $m-1$ $a_{i'}$'s. Observe that there are $m-k$ copies of $\Pr_j(R)$ for all $R$ where $a_i$ beats exactly $m-k$ other alternatives. We can distribute the $m-k$ copies to the $m-k$ bins (one in each) for all $a_{i'}$'s that are ranked after $a_i$. We do this for all possible rankings and in the end, each bin labeled by $a_i\succ a_{i'}$ gets the probabilities of all rankings compatible with $a_i\succ a_{i'}$. This finishes the proof.
\end{proof}

We note that for the Plackett-Luce model, $p^{a_i}_j$'s and $p^{a_i\succ a_{i'}}_j$'s are easy to compute (see Section~\ref{sec:pl}). 
\begin{figure*}[htp]
	\centering
	\includegraphics[width=0.45\textwidth]{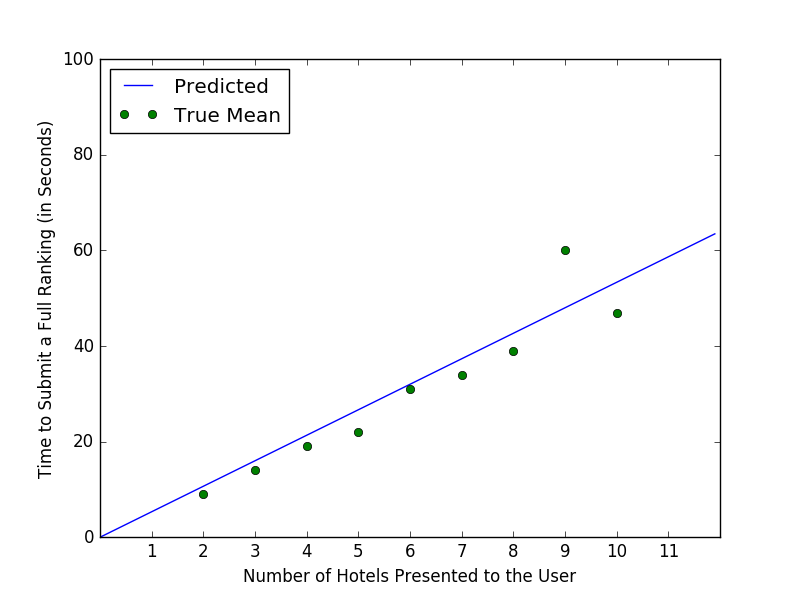}\includegraphics[width=0.45\textwidth]{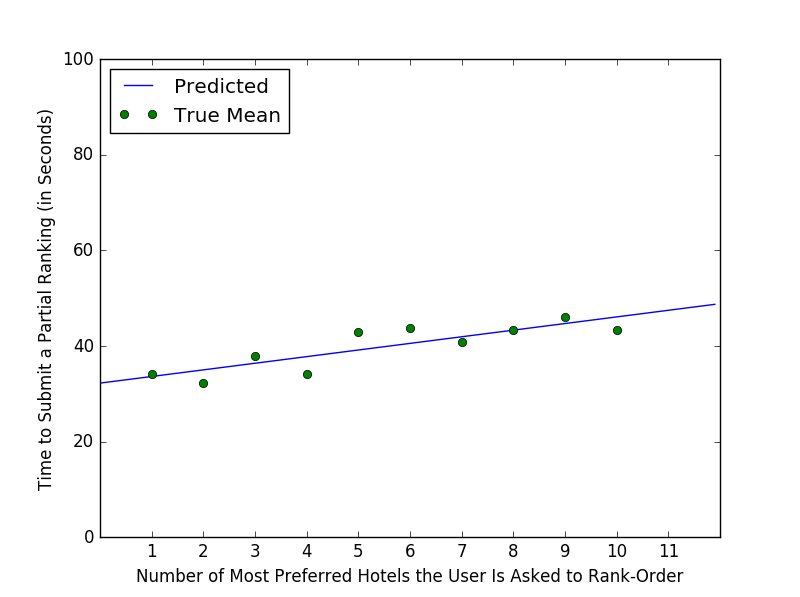}
	\caption{The left subfigure shows the average time a user spent to submit a full ranking over $2, \ldots, 10$ alternatives; the right subfigure shows the average time a user spent to give her ranked top $1, \ldots, 10$ alternatives when 10 alternatives were proposed.}
	\label{fig:time_ranking}
\end{figure*}

\section{EXPERIMENTS}\label{sec:exp}

We first introduce an example of empirically estimating the cost of asking different types of questions on MTurk. Then, we show the result of a simulation of cost-effective preference elicitation using synthetic data.

\begin{figure}[H]
	\centering
	\includegraphics[width=0.4\textwidth]{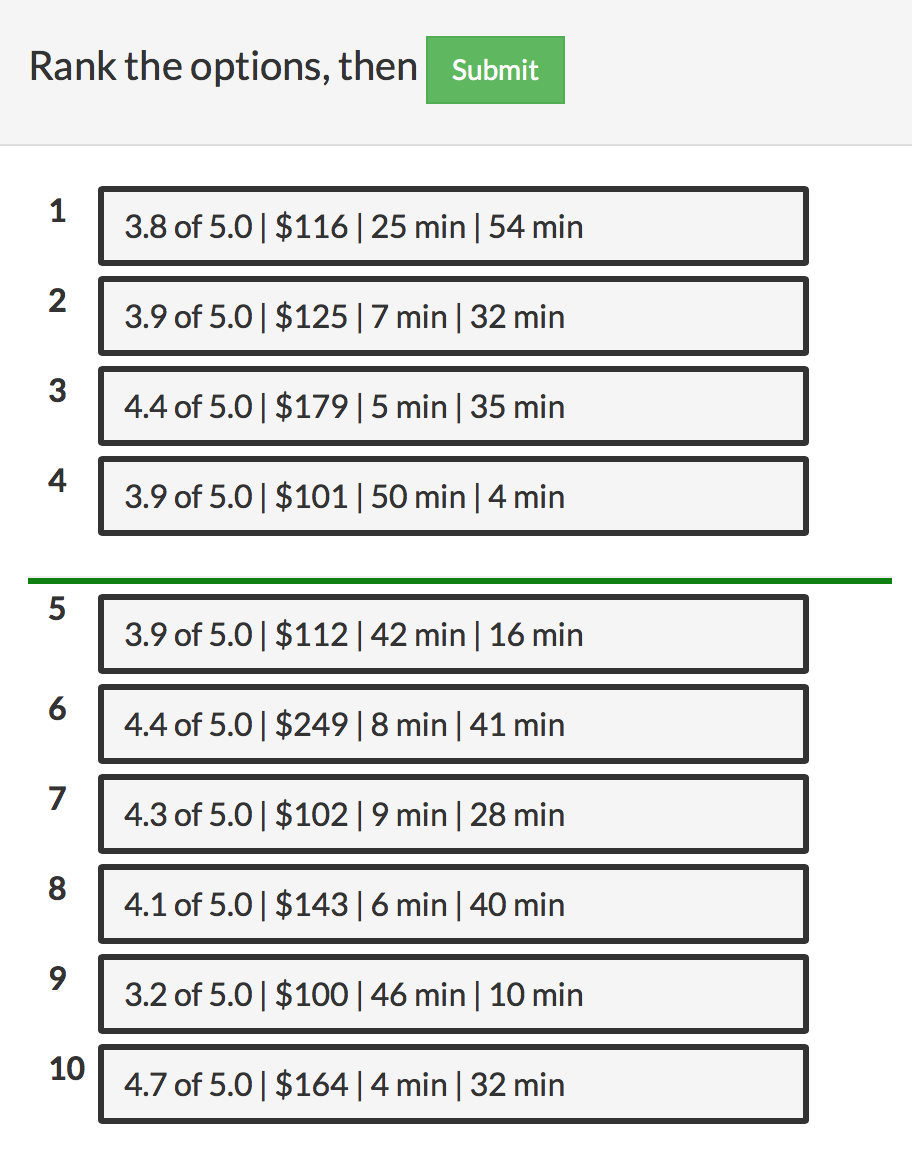}
	\caption{The user interface for a Turker to submit her ranked top 4 over 10 alternatives. The attributes are average ratings, prices per night, time to Times Square, and time to the nearest airport.}
	\label{fig:partialUI}
\end{figure}

\subsection{ESTIMATING $w(h)$}\label{sec:mturk}
We recall that a question is defined by a pair of parameters $(k,l)$, where $l$ is the number of alternatives that are presented to an agent and $k$ is the number of alternatives that the agent is asked to rank at the top $k$ positions.

In order to map the question types to the time to answer them, we run 2 experiments with multiple tasks on MTurk. Each task required MTurk workers to report their preferences over a set of hotels. We recorded the time they spent on each task, in order to learn such mapping in the following two cases:\\
$\bullet$ $k,l \in [2,10], k=l-1$: full rankings;\\
$\bullet$ $k\in [1,10], l = 10$: ranked top $k$ alternatives over 10.

{\bf Experiment Setting.} For the first case, we looked for information on the first 54 Hotels in New York City in alphabetical order. We split the 54 randomly into 9 sets, each containing 2, 3, ..., 10 hotels. We then showed the 9 sets to MTurk workers, randomizing the order of the 9 sets as well as the initial display order of alternatives within each set, and asked them to rearrange by drag-and-dropping the alternatives according to their preferences. The alternatives were anonymous and represented by 4 attributes: average guest rating on a popular travel website, price per night, time to Times Square and time to the nearest airport. 

For the second case, a separate experiment is run with another 10 sets of NYC hotels, drawn randomly again from the first 100 hotels in NYC in alphabetical order. In each task, we placed a horizontal green bar underneath the alternative above which are the $k$ alternatives of interest. We instructed the MTurk workers before the experiment started that only the alternatives above the green bar would count, i.e. only needed to rank-order top-$k$. All of these were done with goal of minimizing the overhead time for workers to understand the instruction so that the recorded time accurately reflect the time of decision-making. An example of the UI is show in {Figure~\ref{fig:partialUI}}, where we asked Turkers to rank-order her top-4 favorite hotels over a set of 10. 

The following analysis is made possible by responses from 408 MTurk workers (202 for the first case and 206 for the second).

{\bf Experiment Results.} Although \cite{Volkov14:Optimal} considered both linear and quadratic cost function and argued for the superiority of the latter, for simplicity, we perform a linear regression on the dataset to obtain a linear cost function. We regress the time to rearrange the alternatives and submit a full ranking on the number of alternatives in the set. 
We find that, on average, the time a Turker spent on rank-ordering a full ranking over $l$ alternatives is $t_{\text{full}-l}=5.33l$ (Figure~\ref{fig:time_ranking} left), and that on rank-ordering her top-$k$ alternatives over 10 alternatives is $t_{\text{top-}k}=1.38k+32.25$ (Figure~\ref{fig:time_ranking} right). In addition, the 408 workers spent an average of $341.5$ seconds on the tasks and were each paid $\$0.3$. Therefore, the monetary cost of elicitation on average is $w = 0.00088t$, which correspond to an hourly wage of $\$3.16$.


Combining these two functions with the hourly wage, we propose the following cost functions, which estimates the cost (in USD) of elicitation about hotel preferences given 4 alternative attributes: $w_{\text{full}-l} = 0.0047l$ and $w_{\text{top}-k} = 0.0012k+0.028$. We observe that the time a user spent is not very sensitive to $k$. This is sensible, as when a MTurk worker ranks her top $k$ choices, she may follow the following procedure: 1.~read the descriptions of all hotels, 2.~form their preferences, and 3~choose top $k$. Step 1 and 2 do not depend on $k$ and dominates the time for step 3, as illustrated in the right figure of Figure~\ref{fig:time_ranking}.
This suggests that when a fixed number of alternatives is proposed to an agent, it's likely that the most cost-effective question to ask is a full ranking, as we will see in the next subsection.

\subsection{COST-EFFECTIVE PREFERENCE ELICITATION}

We demonstrate the viability of our cost-effective framework and compare performances of different information criteria on synthetic data. 

\begin{figure*}[htp]
	\centering
	\includegraphics[width=0.45\textwidth]{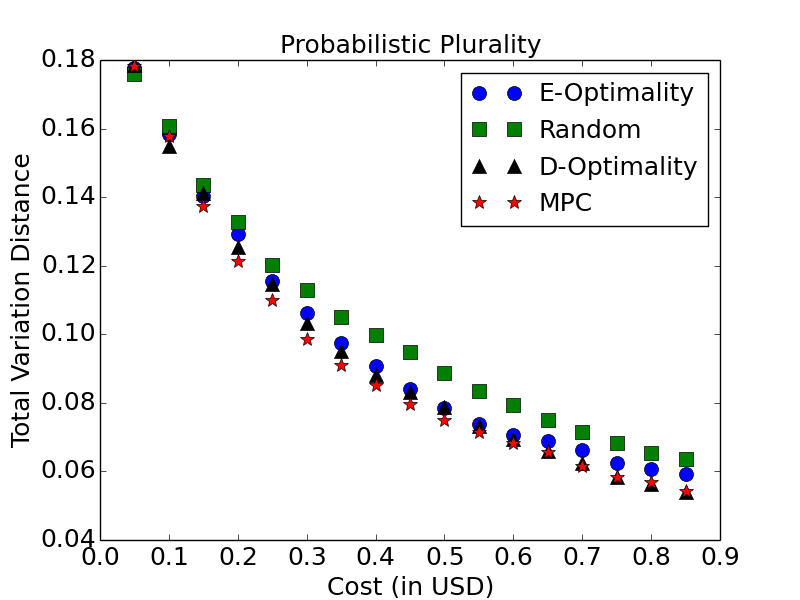}\includegraphics[width=0.45\textwidth]{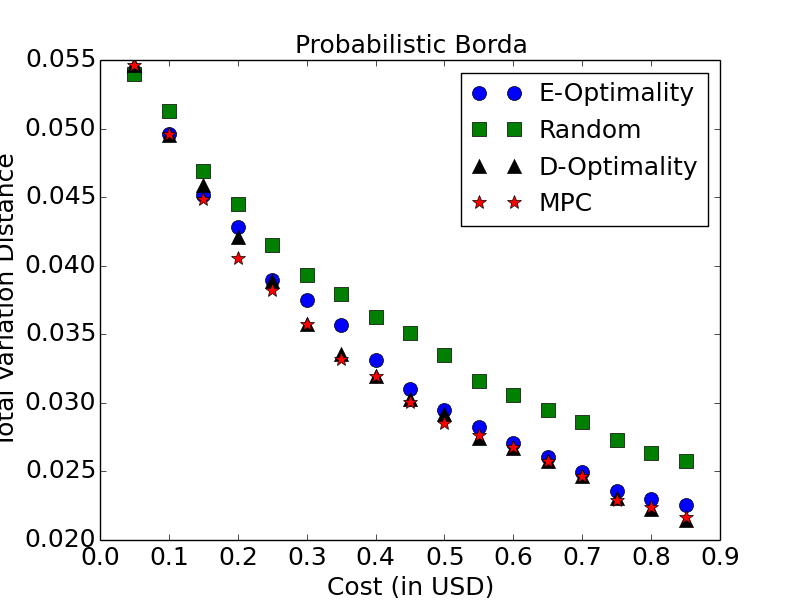}
	\caption{Total variation distance for probabilistic plurality (left) and probabilistic Borda (right).}
	\label{fig:5key_pnb}
\end{figure*}

{\bf Synthetic Data.} We randomly generated 10 alternatives, each of which has $3$ attributes, independently normally distributed $N(0, 1)$. We then randomly generated 5 agents that forms the key group and 20 the regular group. Each agent also has $3$ attributes, independently normally distributed $N(0, 1)$. $B$ was generated from Dirichlet distribution Dir($\vec 1$). The result is averaged over 400 trials.

To echo the motivating example from the beginning of this paper, we simulated the process of eliciting key group's preference by asking agents in the regular group questions. For simplicity, we consider 3 types of questions, represented in $(k,l)$: $(1,2)$, $(1,10)$ and $(9,10)$. We run Algorithm 1 using 3 different information criteria: D-Optimality, E-Optimality, and the proposed MPC. The three elicitation processes utilized the cost function estimated in Section 5.1. They were initialized with the same set of 50 randomly generated pairwise comparisons and was given a \$0.9 budget. Agents' answers to the elicitation questions are generated from the Plackett-Luce model. 

\begin{figure*}[h!]
	\centering
	\includegraphics[width=0.45\textwidth]{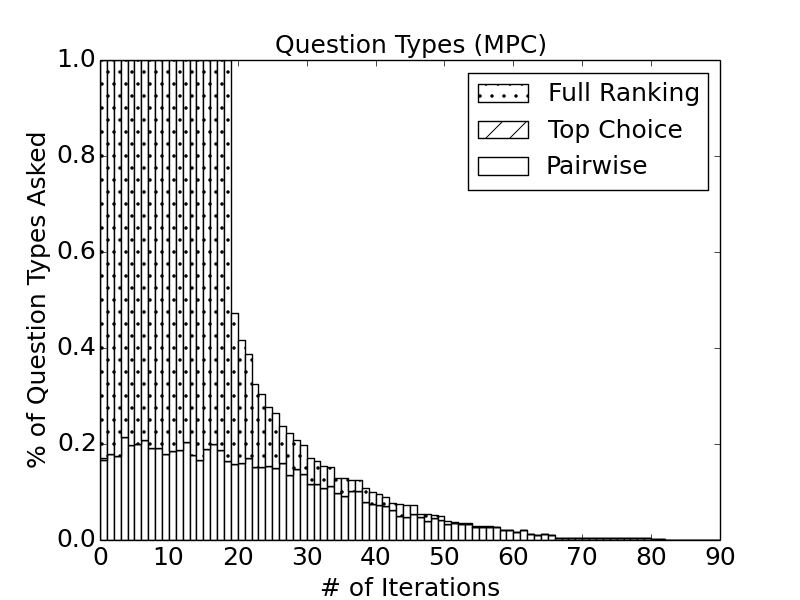}\includegraphics[width=0.45\textwidth]{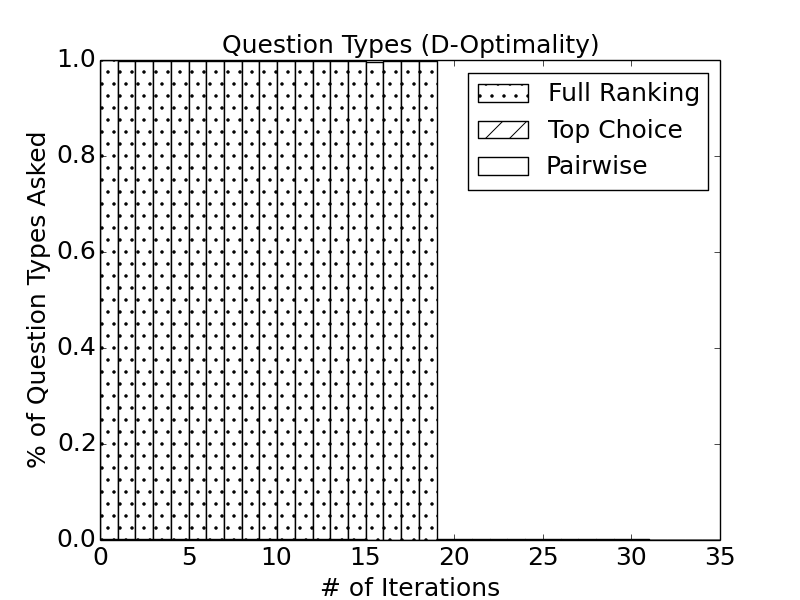}
	\caption{Types of questions chosen by the MPC (left) and D-optimality (right). The legend ``Full Ranking", ``Top Choice", and ``Pairwise" correspond to $(k=9, l=10)$, $(k=1, l=10)$ and $(k=1, l=2)$ respectively.}
	\label{fig:5key_qtype}
\end{figure*}

{\bf Metrics.} We use total variation distance to measure the difference between the winner distributions computed from the ground truth parameter and the estimates, denoted by $\psi^*$ and $\psi$ respectively. The total variance distance is defined as
$\delta(\psi^*, \psi) = \frac 1 2\sum^m_{i=1}|\psi^*(a_i)-\psi(a_i)|$.
To plot the results, at each cost $w$, we used the data point that is below $w$ but closest to $w$ in each trial. These points were averaged over all trials. 

{\bf Observations.} 
We observe that for both probabilistic plurality and probabilistic Borda, the performances of MPC, D-optimality, and E-optimality are similar, all of which significantly outperforms random elicitation questions (see Figures~\ref{fig:5key_pnb}). For example, for probabilistic plurality and probabilistic Borda, at the budget of $0.85$  dollars, MPC achieves 15\% less total variation distance than that of random elicitation questions. As another example, to achieve the total variation distance of 0.064 under randomized plurality (respectively, randomized Borda), MPC uses 20\%  (23.5\%) less money than that of random elicitation questions.

We also observe that D-optimality almost always choose a full ranking as the most cost-effective question, while MPC tends to choose more full rankings than pairwise comparisons at early stages (see Figure~\ref{fig:5key_qtype}). Due to the budget limit, many trials finish after 19 iterations because they only query full rankings. Others finish at different iterations. The distribution of types of questions for E-optimality is similar to MPC. Under all criteria except random, $l=10, k=1$ questions were rarely asked. 

{\bf Discussions.} The meaning of cost-effectiveness in this paper is twofold: (1) in the preference elicitation procedure, we ask elicitation questions that is expected to provide more information per unit cost; and (2) the presence of regular group gives us a belief on the key group's preferences inexpensively. As we have seen in our experiments, a budget of \$0.9 gives us a reasonably good estimate of the key group's preferences by querying the regular group.

\section{CONCLUSIONS AND FUTURE WORK}

We proposed a flexible and cost-effective framework for preference elicitation that can be adapted for any ranking model, any information criterion, and any set of questions. We used randomized voting rules to help make group decisions and proposed MPC for both prediction of one agent's preference and aggregation of a group of agents' preferences. Experiments show that MPC and other commonly-used information criteria work better than asking random elicitation questions. For future work we will explore better information criteria for group decisions. We also plan to extend this framework to multiple types of resource constraints.

\subsubsection*{Acknowledgements}
We thank all anonymous reviewers for helpful comments
and suggestions. This work is supported by NSF \#1453542
and ONR \#N00014-17-1-2621.

\bibliographystyle{plainnat}
\bibliography{references}

\end{document}